\documentclass[runningheads,11pt]{llncs}
\usepackage{appendix}
\usepackage{array}
\usepackage{graphicx}
\usepackage{lipsum}
\usepackage{color, soul}
\usepackage{hhline}
\usepackage{amsmath}
\usepackage{graphicx}
\usepackage{graphics}
\usepackage{enumerate}
\usepackage[export]{adjustbox}
\usepackage{tipa}
\usepackage{adjustbox}
\usepackage{tabularx}
\usepackage[colorlinks=True,citecolor=blue]{hyperref}
\makeatletter
\newcommand{\printfnsymbol}[1]{%
  \textsuperscript{\@fnsymbol{#1}}%
}
\makeatother

\newcolumntype{P}[1]{>{\centering\arraybackslash}p{#1}}

\begin{document}
%

\title{Using Program Synthesis and Inductive Logic Programming to solve Bongard Problems
}
\titlerunning{Using Inductive Programming to solve Bongard Problems}

\author{Atharv Sonwane\inst{1}\thanks{equal contribution}  \and
Sharad Chitlangia\inst{1}\printfnsymbol{1} \and
Tirtharaj Dash\inst{1} \and
Lovekesh Vig\inst{2} \and
Gautam Shroff\inst{1,2} \and
Ashwin Srinivasan\inst{1}}

\authorrunning{Atharv Sonwane et al.}

\institute{APPCAIR, BITS Pilani, K K Birla Goa Campus \and
TCS Research, New Delhi\\}

\maketitle              
\begin{abstract}
The ability to recognise and make analogies is often used as a measure or test of human intelligence \cite{hofstadter1995fluid}. The ability to solve Bongard problems is an example of such a test \cite{hofstadter1995fluid}. It has also been postulated that the ability to rapidly construct novel abstractions is critical to being able to solve analogical problems \cite{Chollet2019OnTM,mitchell2019artificial}. Given an image, the ability to construct a program that would generate that image is one form of abstraction, as exemplified in the Dreamcoder project \cite{Ellis2021DreamCoderBI}. In this paper, we present a preliminary examination of whether programs constructed by Dreamcoder can be used for analogical reasoning to solve certain Bongard problems. We use Dreamcoder to discover programs that generate the images in a Bongard problem and represent each of these as a sequence of state transitions. We decorate the states using positional information in an automated manner and then encode the resulting sequence into logical facts in Prolog. We use inductive logic programming (ILP), to learn an (interpretable) theory for the abstract concept involved in an instance of a Bongard problem. Experiments on synthetically created Bongard problems for concepts such as ‘above/below’ and ‘clockwise/counterclockwise’ demonstrate that our end-to-end system can solve such problems. We study the importance and completeness of each component of our approach, highlighting its current limitations and pointing to directions for improvement in our formulation as well as in elements of any Dreamcoder-like program synthesis system used for such an approach.


\end{abstract}
\section{Introduction}
It has long been understood that choice of representation can make a significant difference
to the efficacy of machine-based induction.
``An understanding of the relationship between problem formulation and problem-solving efficiency is
a prerequisite for the design of procedures that can automatically choose the most `appropriate'
representation of a problem (they can find a `point of view' of the problem that maximally
simplifies the process of finding a solution'', Amarel \cite{amarel1981representations}. In fact, in the previous quote, by `choose' what
is probably meant is `construct', if we are to avoid kicking Feigenbaum's bottleneck down the
road from extracting models to extracting representations. However, quoting Melanie Mitchell in 2021 \cite{mitchell2019artificial}
``...enabling machines to form human-like conceptual abstractions is still an almost completely unsolved problem.’’

\begin{figure*}[!t]
    \centering
    \includegraphics[scale=0.6]{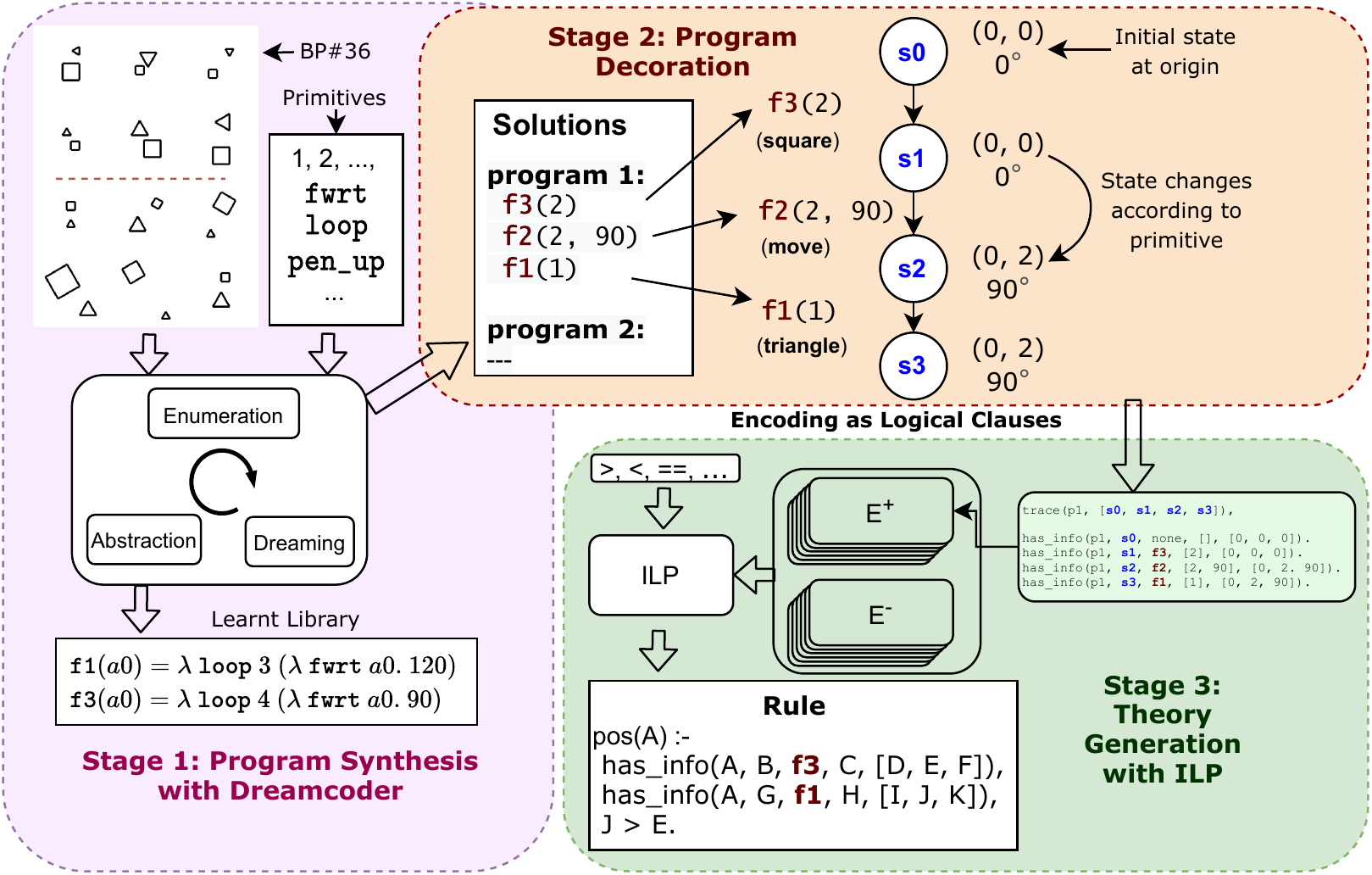}
    \caption{The proposed Inductive Programming System}
    \label{fig:overall}
\end{figure*}


In 1967, Mikhail Bongard proposed \cite{Bongard2018PatternR} a set of 100 puzzles, ordered by increasing complexity,
designed to challenge the abstraction and analogy-making abilities
of both humans and AI systems \footnote{As explained in \cite{mitchell2019artificial}: (i) constructing abstractions to draw similarities between observations constitutes an analogy (ii) analogies that occur often become categories/classes/concepts (e.g. fruit, `roundness', `theft') and are no longer referred to as analogies (iii) thus the term analogy is most often used for situations requiring novel abstractions, e.g. in few-shot scenarios such as Bongard problems.}. 
Each Bongard problem captures a unique abstract concept in the space of drawings
of shapes and lines with six positive and six negative examples. 
Since the concepts required for each Bongard problem are different, an AI
system for solving them would need to form the required (novel) abstractions 
from very few examples (a feature of `analogical' reasoning \cite{mitchell2019artificial}),
using simpler abstractions that are both sufficiently powerful as well as problem-specific. Further, since there are only a hundred or so Bongard problems, a 
data-intensive approach that assumes access to large volumes of similar problems is ruled out \footnote{Quoting \cite{mitchell2019artificial}:
``In short, today’s ConvNets, . . . do not have what it takes to do the kinds of abstraction ... required even in Bongard’s idealized problems’’}.



Thus the question arises as to what kind of representations might suffice to enable
rapid analogy-making, i.e., capturing the `essential’ similarity between one set of drawings
vs another.

`Dreamcoder’, a recently proposed program induction system \cite{Ellis2021DreamCoderBI} synthesizes a \textit{program} to solve 
input/output problems by creating useful abstractions across problems. From starting
with very simple primitives such as `move forward’, `pen up/down’, etc., Dreamcoder
can \textit{abduce} new functions such as `draw a triangle’ to draw more complex figures.





There are good reasons to
look at this form of program-synthesis as a mechanism for automated representation learning:
(a) Empirical results with programs like Dreamcoder show that it is possible to synthesize
	programs for diverse tasks from a very small number or even a single example;
(b) The Symbolic language adopted for primitive functions ($\lambda$-expressions) 
    is sufficiently expressive for constructing programs
	for data of any type; 
(c) The intermediate representations have clearly defined interpretations, based on functional
	composition. 
	Further, Inductive Logic Programming (ILP~\cite{Muggleton1994InductiveLP})
	can use these symbolic abstractions for constructing
	human-readable discriminatory models.

We postulate that programs synthesized to re-draw the figures in a Bongard problem should serve
as good intermediate abstractions from which to learn, `on the fly’, the concept involved.
For example, if the concept were `sameness’, programs that generated positive drawings would contain repeated use of the same higher level functions.
We also find there are some high-complexity Bongard problems for which program synthesis fails, 
as well as cases where the program as a representation alone may 
be insufficient;
and we postulate possible mechanisms to alleviate these difficulties
in the future.

\vspace{-10pt}

\section{Related Work}
Recent critical reviews \cite{Chollet2019OnTM,mitchell2019artificial} of progress in AI research, posit that few-shot visual reasoning is one of the hallmarks of human intelligence and more than half a century of research in AI, progress on that end is still limited. 
Bongard Problems~\cite{Bongard2018PatternR} are a popular set of 
visual reasoning puzzles for testing AI systems.
To the best of our knowledge, the maximum number of problems solved by an artificial intelligence system is 42 \cite{Depeweg2018SolvingBP}. Similar to our approach, the approaches proposed in \cite{Depeweg2018SolvingBP,wong2021leveraging} uses a visual language and a pragmatic approach to solve Bongard Problems. In our approach, the visual language is composed of graphical $\lambda$-expression programs that are rich in abstractions. Hence, the novelty in our approach is the amount of background knowledge encoded in the graphical language. For instance, even simple shapes such as triangles, squares, circles are invented before being used for more complicated visual reasoning tasks.

Michie's original definition of Machine Learning was in terms of two orthogonal axes - predictive accuracy and comprehensibility of hypotheses. Where approximate methods such as Neural Networks and Support Vector Machines might excel on predictive accuracy, symbolic methods such as Inductive Programming excel at being comprehensible. In addition to being interpretable, programs, in particular logic programs, have been known to show strong generalization capabilities \cite{Silver2019FewShotBI}. Through our 3-staged approach that solves some Bongard problems we show the benefits of using programs as representations. 


\section{Methodology}
Our 3-staged inductive programming approach to solving Bongard Problems is shown in Fig.~\ref{fig:overall} and described below:


\noindent 
\textbf{Dreamcoder.}
The input to Dreamcoder is a set of tasks $X$, and the goal is to both infer a program $\rho_x$ solving each task $x \in X$ and learn a library $L$ of primitives and program abstractions encoding a prior distribution $P(\rho | L)$ that will enable it to solve tasks in that domain. In our system, Synthesized $\lambda$-expressions from Dreamcoder act as a generative representations of images in a Bongard Problem. Given the set of 12 images of a Bongard problem, Dreamcoder outputs 12 programs, one for generating each image.

\noindent
\textbf{Transducer.} This module converts the synthesised programs into a first order logic representation based the program's decorated state transition diagram. 

$\lambda$-expression programs that generate images of a Bongard problem in our system are purely functional. Functional Programs are stateless in nature. But as noted originally in Dreamcoder \cite{Ellis2021DreamCoderBI}, LOGO-style graphics programs are implemented using a state monad and by encoding each primitive action in a continuation passing style. This enables representation of the purely functional program in an imperative-style state transition diagram.

Imperative programs can be seen as sequences of instructions (primitives) each of which when executed cause a \textit{transition} from one \textit{state} to another. We hypothesise that these states and primitive calls in programs representing images from a Bongard problem hold valuable information for identifying its differentiating concept. The state information is extracted from the result after evaluating sub-programs and is used to \textit{decorate} a state-transition diagram of the overall program (see Figure \ref{fig:overall}).  In the current implementation of the system, we limit our program state to the very basic type of information that may be extracted: positional coordinates (x,y) and orientation of the cursor (on the drawing canvas). More complicated state information can also be added using feature extractors (such as the ones used in \cite{Depeweg2018SolvingBP}) working on top of the partial program or the partially generated image. For instance, to extract high level perceptual information such as if two shapes are touching or not, it might be difficult to extract directly from a sub-program but can be achieved by means of an image processing module working on top of the graphic drawings of the produced programs. 


The transducer generates the decorated state transition diagram for a given program as described above and represents this as a set of FOL statements. Here, the sequences of states 
for some program {\tt p1} are specified by the {\tt trace/2} predicate as:
{\tt trace(p1,[s0,s1,s2])}.
Then each state in the trace is associated with its decorated information
by representing using a {\tt has\_info/5} predicate. For example the fact that the state {\tt s0} of program {\tt p1}
uses primitive {\tt f0} with arguments
{\tt [1,2]} after which the co-ordinates become {\tt (2.0,2.0)}
with orientation {\tt 1.33} can be written as:
{\tt
has\_info(+Program,-State,\#Primitive,-Args,[-X,-Y,-Angle])}.


\noindent
\textbf{ILP.}
ILP has been the focus of much research to solve Bongard problems \cite{10.5555/242040.242063} due to the logical nature of the puzzles and its formulation of learning a theory that differentiates, using some background knowledge, positive from negative examples. We use Aleph \cite{srinivasan2001aleph} to learn a classification theory where the background knowledge for positive and negative examples are the set of FOL facts output from transducer corresponding to the generated programs for the positive and negative images in a Bongard problem respectively. The BK is encoded using the {\tt trace/2} and {\tt has\_info/5} predicates as described previously.

%
%

\vspace{-10pt}
\section{Empirical Evaluation}
We evaluate our system on a set of synthetic Bongard problems with the following aims: (1) Can our end-to-end system solve a reasonable set of Bongard problems? (2) What are some of the breaking points and limitations of our 3-staged inductive programming system in the context of solving Bongard problems and more generally, for visual reasoning?

%
\vspace{-10pt}
\subsection{Materials}
\textbf{Data.}
We synthetically create a set of 14 Bongard problems each consisting of 12 images (6 positive,
6 negative)\footnote{\label{foot:url}These are available at this \href{https://drive.google.com/drive/folders/1Uw_9n7HQd2C1AR62oTwPVhJLOJFP85LV?usp=sharing}{link}.} that match original concepts encoded in
Bongard problems.
We aim for a representative sample of Bongard problems from recent studies. For instance we include problems (e.g. BP~$\#21$ and BP~$\#60$) for which a convolutional neural network\cite{stabinger20165,yun2020deeper} fails to learn the concept encoded.


\noindent
\textbf{Background Knowledge.}
The provided primitives to Dreamcoder (see \ref{foot:url}) serve as its background knowledge. For the ILP stage, we provide additional comparison style predicates such as \verb+==+, \verb+\==+ for comparison of arguments, \verb+>+, \verb+<+, for comparison of {\tt (X,Y)} coordinates, \verb+bw_90_270+, \verb+lt(90)+, etc., for comparison of orientation.


\subsection{Results}

The results for various bongard problems, including the libraries learnt with Dreamcoder, the final theories learnt with ILP and their explanations, are shown in Fig.~\ref{fig:result-pos-theories}. Our system is able to solve 8 of the total 14 problems considered. On the easier end the tasks solved include basic visual reasoning concepts such as the presence of circle (BP $\#24$) and clockwise spirals (BP $\#16$). These are solved by simply learning a theory to check for the relevant invented primitive in the $\lambda$-expression program. On the more complicated end, where some of the current systems fail \cite{stabinger20165} are BP $\#21$, BP $\#23$, BP $\#36$, BP $\#60$, BP $\#75$. Notably, in \cite{stabinger20165} it was found that even with $20,000$ examples, a CNN could not learn the high-level concept encoded in BP $\#21$ and BP $\#60$. 


\vspace{-1pt}
\noindent
\textbf{Improving Graphical Program Synthesis}. 
The two major limitations of our current system on the graphical program synthesis module can be seen in Fig.~\ref{fig:result-neg-theories}: (1) It cannot directly work on top of hand-drawn images, due to lack of certain primitives, such as BP $\# 4$, BP $\# 14$, BP $\# 94$  and (2) It cannot draw figures where there might be a lot to draw.
Apart from the inclusion of the required primitives, it may be possible to get around (1) by comparison of the output of a graphics program with hand-drawn images using a learned metric as previously done in \cite{ellis2017earning}. 
For (2), the enumerative search itself might need to be guided with more than simply a distribution over primitives as previously demonstrated in the sub-area of execution guided synthesis such as \cite{ellis2019rite} which works by evaluating partial programs to direct the search early-on.

\vspace{-1pt}
\noindent
\textbf{Improving State Decorations}. 
The state decorations can be augmented with learned features
from Dreamcoder-produced programs or the images. These features could be propositional (e.g. `Is the first argument of last executed primitive greater than 2?'). It is interesting to note here that a wide variety of  feature extractors could be utilised to add to the state decorations., including those described in \cite{Bongard2018PatternR} which was able to solve 42 Bongard Problems when evaluated exhaustively.


\noindent
\textbf{Improving the final theory learning step}.
It is of interest to investigate if multiple theories learned across different Bongard puzzles can be used for inventing new clauses
via inverse-resolution techniques (such as inter-construction) that might serve to improve theory learning for subsequent puzzles:
For example, the concept of `smallness' (as seen in the theories for BP~\#21 and BP~\#53 where it is expressed as the length of sides being less than number of sides for a shape) over multiple problems can
be generalised by a meta-rule allowing learning of simpler discriminative theories.
It may also be possible to construct meta-rules that generalise over sub-programs. If we assume (sub-)programs are computational encodings of predicates, then such meta-rules would be statements in at least 2nd order logic, that allow quantification over predicates. 

\begin{figure}[!htb]
\begin{adjustbox}{width=\columnwidth,center}
\begin{tabular}{|c|c|c|c|c|c|c|} 
\hline
\textbf{Concept} & \begin{tabular}[c]{@{}l@{}}Concave vs\\Convex (\#4)\end{tabular} & \begin{tabular}[c]{@{}c@{}}Large Total Line \\Length (\#14)\end{tabular} & \begin{tabular}[c]{@{}c@{}}Collinearity\\(\#40)\end{tabular} & \begin{tabular}[c]{@{}c@{}}Number of\\Lines (\#85)\end{tabular} & \begin{tabular}[c]{@{}c@{}}Location \\of solid \\shape (\#94)\end{tabular} & \begin{tabular}[c]{@{}c@{}}Nuber of~\\sides of \\hatching (\#96)\end{tabular}  \\ 
\hline
\begin{tabular}[c]{@{}c@{}}\textbf{Failure}\\\textbf{Stage}\end{tabular} & Representation & Representation & Search & Search & Representation & Search \\ 
\hline
\textbf{Explanation} & \begin{tabular}[c]{@{}c@{}}Solid fills cannot\\be represented\end{tabular} & \begin{tabular}[c]{@{}c@{}}Arbitrary curves~\\cannot be\\represented\end{tabular} & \begin{tabular}[c]{@{}c@{}}Two many \\shapes to\\be drawn\end{tabular} & \begin{tabular}[c]{@{}c@{}}Unable to~\\draw 5\\lirregular ines~\end{tabular} & \begin{tabular}[c]{@{}c@{}}Solid fills cannot\\be represented\end{tabular} & \begin{tabular}[c]{@{}c@{}}Too many \\lines to\\be drawn\end{tabular}  \\
\hline
\end{tabular}
\end{adjustbox}
\caption{The failure modes of the Inductive Programming System}
\label{fig:result-neg-theories}
\end{figure}
\begin{figure}[!htb]
\centering
{\renewcommand{\arraystretch}{1.5}
{\tiny{
\begin{tabular}{|p{1.3cm}|p{3cm}|p{3.5cm}|p{3.7cm}|}
\hline
\multicolumn{1}{|c|}{\textbf{Concept}} & \multicolumn{1}{c|}{\textbf{Invented Primitives}} & \multicolumn{1}{c|}{\textbf{Theory}} & \multicolumn{1}{c|}{\textbf{Explanation}} \\
\hline

Anti-clockwise vs Clockwise (BP~\#16) &

    {\tt f2(a0)}, {\tt f3(a0)} both draw anti-clockwise spirals with {\tt a0} controlling tightness of the spiral. {\tt f2} and {\tt f3} use step length 2 and 1. &
    
    {\tt pos(A):- has\_info(A,B,f3,C,[D,E,F]).} 
    \vspace{-0.1cm}
    
    {\tt pos(A):- has\_info(A,B,f2,C,[D,E,F]).} &
    Presence of invented primitive for drawing spirals that are anti-clockwise. \\ \hline
    
Smaller shape present (BP~\#21) &

    {\tt f1(a0, a1)}: Draw an {\tt a0}-sided polygon with sides of length {\tt a1}  &
    
    {\tt pos(A): -}
        {\tt has\_info(A,B,rtfwint,C,[D,E,F]),}
        {\tt C=[G|H],H=[I|J],G>I},
        {\tt has\_info(A,K,f1,L,[D,E,F]).} 
    \vspace{-0.1cm}
    
    {\tt pos(A):-}
        {\tt has\_info(A,B,f1,C,[D,E,F]),}
        {\tt C=[G|H],H=[I|J],G>I.} & 
    
    Program contains a move primitive where the division factor for angle is greater than multiplication factor for distance, \hfill \textit{or} \hfill there is a polygon with side length less than number of sides. Indicating the shape is small. \\ \hline

Number of Shapes (BP~\#23) & 
None & 
{\tt pos(A):-}
{\tt trace(A,B),B=[C|D],D=[E|F],F=[]} 
& Contains only 2 states (in which only one shape can be drawn since first state is always initial state). \\
\hline

Circle present (BP~\#24) & 
{\tt f0(a0)}: Draw a circle with radius {\tt a0} &
{\tt pos(A):-}  {\tt has\_info(A,B,f0,D,[E,F,G]).} & 
Presence of invented primitive for drawing circle. \\
\hline

Triangle above Square (BP~\#36) &
{\tt f1(a0)}: Draws a triangle of side length {\tt a0}. 
{\tt f3(a0)}: Draws a square of side length {\tt a0} &
{\tt pos(A):-} {\tt has\_info(A,B,f3,C,[D,E,F]),} {\tt has\_info(A,G,f1,H,[I,J,K]),} {\tt J>E.}
& 
Triangle exists with y coordinate greater than that of square\\
\hline

Enclosed shape has fewer sides (BP~\#53) &
{\tt f1(a0, a1)}: Draw an {\tt a0}-sided polygon with sides of length {\tt a1} &
{\tt pos(A):-}
{\tt has\_info(A,B,f0,C,[D,E,F]),}
{\tt has\_info(A,R,pt,Q,[K,L,M]),}
{\tt has\_info(A,I,f0,J,[K,L,M]),}
{\tt C=[G|H],J=[N|O],O=[P|Q],}
{\tt G>N, N>P.} &
Smaller polygon (having length of side smaller than number of sides) has has fewer sides than larger (enclosing) polygon. \\
\hline

Two Similar Shapes (BP~\#60) & 
None & 
{\tt pos(A):-} 
{\tt has\_info(A,B,C,D,[E,F,G]),} 
{\tt has\_info(A,H,C,I,[J,K,L]),}
\verb|H\==B.| & 
There are two shapes which are different (i.e. calls on two different invented primitives)  
\\
\hline

Triangle on concave side of arc (BP~\#75) & 
{\tt f0(a0)}: Draws an arc of radius {\tt a0} \hfill {\tt f1(a0)}: draws a triangle of side length {\tt a0} & 
{\tt pos(A):-}
{\tt has\_info(A,B,f0,C,[D,E,F]),}
{\tt bw\_90\_270(F),} 
{\tt has\_info(A,G,f1,H,[I,J,K]),}
{\tt D>I.}
\vspace{-0.1cm}

{\tt pos(A):-} {\tt has\_info(A,G,f1,H,[I,J,F]),}
{\tt bw\_270\_90(K),I>D.} &
Either the arc is facing right hand side and x coordinates of triangle is greater than those of arc, \hfill \textit{or} \hfill arc is facing left and x coordinate of triangle is less than that of arc. \\
\hline
\end{tabular}}}
\caption{Solutions for Bongard Problems using our approach.}
\label{fig:result-pos-theories}}
\end{figure}

\vspace{-12pt}
\section{Conclusion}
Recent progress in program synthesis exemplified by projects such as Dreamcoder \cite{Ellis2021DreamCoderBI} led us to 
examine whether a program generated to re-construct an image might serve as a good representation
for rapidly learning new visual concepts, such as required while solving Bongard problems. We have presented
an end-to-end automated system combining inductive functional program synthesis using Dreamcoder and theory induction
using Inductive Logic Programming and used it to conduct experiments on solving Bongard problems.

Our approach is indeed able to solve certain Bongard problems, as well as produce theories that are interpretable;
using only few examples unlike recent deep-learning approaches. In contrast,
the program synthesis process of Dreamcoder itself involves 'dreaming' examples using a generative mode that aid in discovering 
higher-level programmatic primitives via refactoring that are crucial to enable the downstream ILP learner to
produce a coherent and accurate discriminative theory.

We have also highlighted opportunities for improving both the program synthesis and ILP parts of the process, including possibilities for marrying our approach with deep-learning. We believe our results demonstrate the promise of using automatically synthesised programs as abstract representations  to aid in rapidly learning novel concepts and furthering research in analogical reasoning.
\bibliographystyle{splncs04}
\bibliography{ref}

\begin{thebibliography}{10}
\providecommand{\url}[1]{\texttt{#1}}
\providecommand{\urlprefix}{URL }
\providecommand{\doi}[1]{https://doi.org/#1}

\bibitem{amarel1981representations}
Amarel, S.: On representations of problems of reasoning about actions. In:
  Readings in artificial intelligence, pp. 2--22. Elsevier (1981)

\bibitem{Bongard2018PatternR}
Bongard, M.: Pattern recognition. In: Encyclopedia of Social Network Analysis
  and Mining. 2nd Ed. (2018)

\bibitem{Chollet2019OnTM}
Chollet, F.: On the measure of intelligence. ArXiv  \textbf{abs/1911.01547}
  (2019)

\bibitem{Depeweg2018SolvingBP}
Depeweg, S., Rothkopf, C., J{\"a}kel, F.: Solving bongard problems with a
  visual language and pragmatic reasoning. ArXiv  \textbf{abs/1804.04452}
  (2018)

\bibitem{ellis2019rite}
Ellis, K., Nye, M., Pu, Y., Sosa, F., Tenenbaum, J., Solar-Lezama, A.: Write,
  execute, assess: Program synthesis with a repl. arXiv preprint
  arXiv:1906.04604  (2019)

\bibitem{ellis2017earning}
Ellis, K., Ritchie, D., Solar-Lezama, A., Tenenbaum, J.B.: Learning to infer
  graphics programs from hand-drawn images. arXiv preprint arXiv:1707.09627
  (2017)

\bibitem{Ellis2021DreamCoderBI}
Ellis, K., Wong, C., Nye, M., Sabl{\'e}-Meyer, M., Morales, L., Hewitt, L.B.,
  Cary, L., Solar-Lezama, A., Tenenbaum, J.: Dreamcoder: bootstrapping
  inductive program synthesis with wake-sleep library learning. Proceedings of
  the 42nd ACM SIGPLAN International Conference on Programming Language Design
  and Implementation  (2021)

\bibitem{hofstadter1995fluid}
Hofstadter, D.R.: Fluid concepts and creative analogies: Computer models of the
  fundamental mechanisms of thought. Basic books (1995)

\bibitem{mitchell2019artificial}
Mitchell, M.: Artificial intelligence: A guide for thinking humans. Penguin UK
  (2019)

\bibitem{Muggleton1994InductiveLP}
Muggleton, S., Raedt, L.D.: Inductive logic programming: Theory and methods. J.
  Log. Program.  \textbf{19/20},  629--679 (1994)

\bibitem{10.5555/242040.242063}
Saito, K., Nakano, R.: A Concept Learning Algorithm with Adaptive Search, p.
  347–363. Oxford University Press, Inc., USA (1996)

\bibitem{Silver2019FewShotBI}
Silver, T., Allen, K.R., Lew, A.K., Kaelbling, L., Tenenbaum, J.: Few-shot
  bayesian imitation learning with logic over programs. ArXiv
  \textbf{abs/1904.06317} (2019)

\bibitem{srinivasan2001aleph}
Srinivasan, A.: The aleph manual (2001)

\bibitem{stabinger20165}
Stabinger, S., Rodríguez-Sánchez, A., Piater, J.: 25 years of cnns: Can we
  compare to human abstraction capabilities? arXiv preprint arXiv:1607.08366
  (2016)

\bibitem{wong2021leveraging}
Wong, C., Ellis, K., Tenenbaum, J.B., Andreas, J.: Leveraging language to learn
  program abstractions and search heuristics. arXiv preprint arXiv:2106.11053
  (2021)

\bibitem{yun2020deeper}
Yun, X., Bohn, T.A., Ling, C.X.: A deeper look at bongard problems. In:
  Canadian Conference on AI. pp. 528--539 (2020)

\end{thebibliography}

\end{document}